\documentclass[conference]{IEEEtran}
\IEEEoverridecommandlockouts
\usepackage{amsmath,amssymb,amsfonts}
\usepackage{algorithmic}
\usepackage{graphicx}
\usepackage{textcomp}
\usepackage{xcolor}
\usepackage{color}
\usepackage{arydshln}
\usepackage{booktabs}
\usepackage{subfig}
\usepackage{enumitem}
\usepackage{bm}
\graphicspath{{figures/}}
\usepackage{pgfplots,tikzscale}
\usepackage{balance}
\usepackage{graphbox}

\renewcommand{\baselinestretch}{0.955}

\begin{document}
\title{One-Class Feature Learning Using Intra-Class Splitting}

\author{
	\IEEEauthorblockN{Patrick Schlachter, Yiwen Liao and Bin Yang}
	\IEEEauthorblockA{Institute of Signal Processing and System Theory\\University of Stuttgart, Germany}
}

\maketitle

\begin{abstract}
This paper proposes a novel generic one-class feature learning method based on intra-class splitting. In one-class classification, feature learning is challenging, because only samples of one class are available during training. Hence, state-of-the-art methods require reference multi-class datasets to pretrain feature extractors. In contrast, the proposed method realizes feature learning by splitting the given normal class into typical and atypical normal samples. By introducing closeness loss and dispersion loss, an intra-class joint training procedure between the two subsets after splitting enables the extraction of valuable features for one-class classification. Various experiments on three well-known image classification datasets demonstrate the effectiveness of our method which outperformed other baseline models in average.
\end{abstract}

%

\section{Introduction}

One-class classification is a subfield in machine learning, aiming at identifying normal data from abnormal data using a training dataset consisting merely of samples from the normal class. It is more challenging than binary or multi-class classification in which samples from all classes are available during training. Various one-class classifiers were proposed and successfully applied to a wide range of applications including fault detection, novelty detection or anomaly detection~\cite{Khan2010,Pimentel2014}. Sch\"olkopf et al.~\cite{Schoelkopf2001} or Tax et al.~\cite{Tax2004} proposed state-of-the-art one-class classifiers which are characterized by a tight decision boundary around the training samples of the normal class in order to reject abnormal samples of different kinds during inference.

To achieve a tight decision boundary, one-class classifiers require an input feature space that fulfills the following two conditions. First, features of normal data must be compactly distributed. We call this requirement \emph{closeness}. Second, normal and abnormal data must have large distances between each other in the feature space. This requirement is called \emph{dispersion}. Both requirements are comparable to the linear discriminant analysis in clustering in which the sample distances within a cluster are minimized and the sample distances between clusters are maximized~\cite{Hastie2009}.

As the stated conditions are typically not fulfilled for high-dimensional data such as natural images, the performance of state-of-the-art one-class classifiers is quite limited. Hence, a feature extraction method is necessary for such classifiers to transform raw data into a suitable latent feature space satisfying the closeness and dispersion requirements. However, this is challenging, because abnormal data samples are not available during training. Therefore, few work was done on feature learning for one-class classification.

State-of-the-art feature learning methods for one-class classification are supported by samples from other classes. In particular, a reference dataset is used to pretrain a model on many other classes~\cite{Perera2018}. The underlying assumption is that real abnormal samples might be contained in these classes. However, this assumption is too strong, since the number of other classes is unlimited. If the reference dataset is not representative for real abnormal classes, then the decision boundary of a one-class classifier will be too loose. Therefore, the choice of a reference dataset is crucial. Indeed, a meaningful multi-class reference dataset is typically not available due to the nature of the one-class classification problem.

\begin{figure}
	\centering
	\includegraphics[width=0.6\linewidth,  trim={0.1cm 3.5cm 0.1cm 0.1cm}, clip]{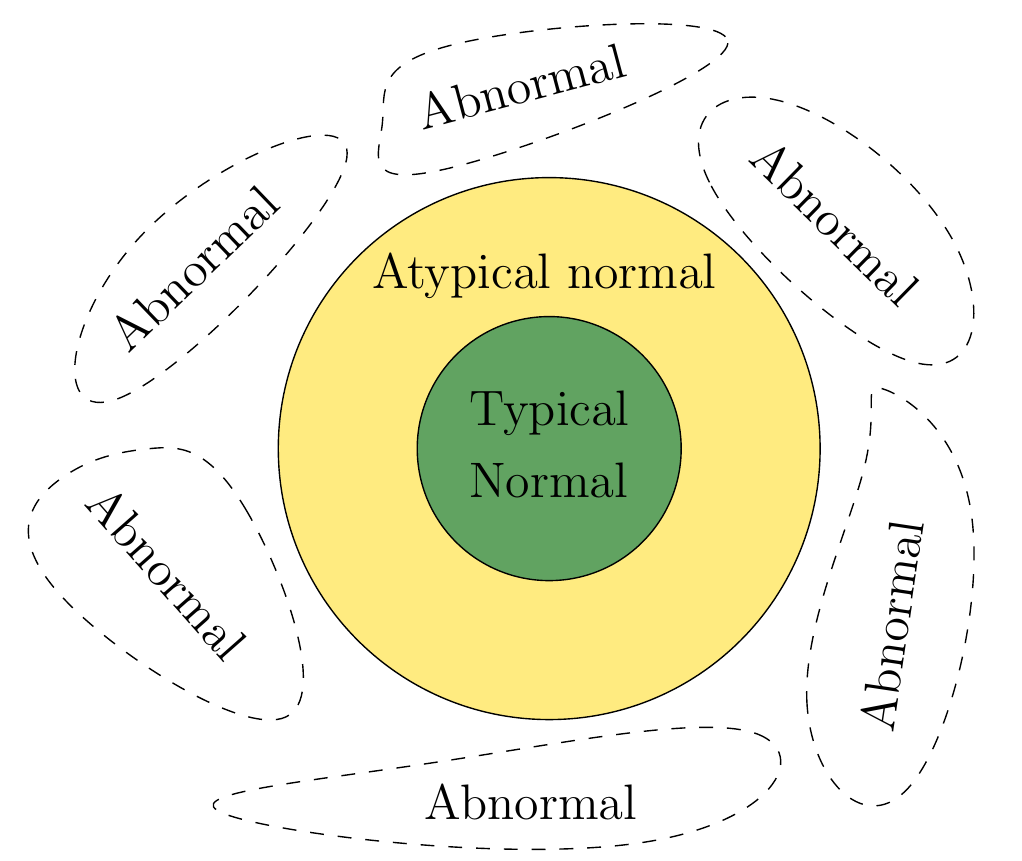}
	\caption[ideavenn]{In a certain latent space, typical normal data (\emph{green}) is clustered and surrounded by atypical normal data (\emph{yellow}). Abnormal data (\emph{dashed lines}) is located outside this area and can be arbitrarily distributed.}
	\label{fig:ideavenn}
	\vspace{-0.5cm}
\end{figure}

In this paper, our goal is to learn a latent space that fulfills both closeness and dispersion requirements using only normal data. As a result, latent representations are optimized to the training class and not biased to any outliers. A solution for this goal does not only enable the use of conventional one-class classifiers to the learned feature space, but has also high potential for unsupervised learning and open set recognition~\cite{Scheirer2013}.
Intuitively, we assume that in a certain latent space, normal data is clustered and abnormal data distributes around it. Fig.~\ref{fig:ideavenn} illustrates this scenario.

In order to find such a latent space, our first idea is to split a normal training set into two subsets in an unsupervised manner and by using a similarity metric. One subset includes typical normal data, meaning the majority of a given training dataset, and the other subset consists of atypical normal data, meaning the minority of the training set. Hence, no reference dataset is necessary in contrast to~\cite{Perera2018}.
By using an joint training procedure to maximize the dispersion between the typical and atypical normal data, a tight decision boundary around the normal data in the latent feature space can be achieved. This is our second key idea.

\begin{figure}
	\centering
	\includegraphics[width=0.49\linewidth, trim={0.5cm 0.5cm 0.5cm 0.5cm}, clip]{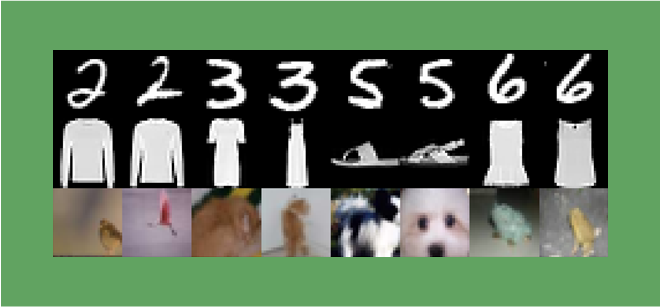}
	\hfill
	\includegraphics[width=0.49\linewidth, trim={0.5cm 0.5cm 0.5cm 0.5cm}, clip]{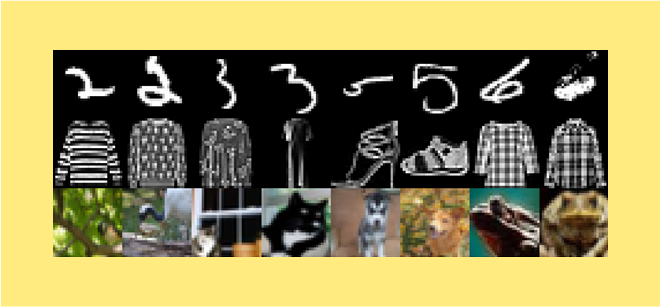}
	\caption[normal and atypical normal data according to SSIM]{\emph{Green}: Exemplary typical normal samples with a higher SSIM score compared to their reconstructions. \emph{Yellow}: Exemplary atypical normal samples having a lower SSIM score compared to their reconstructions.
	The atypical normal samples are more difficult to recognize or have more redundant details compared to typical normal ones.
	}
	\label{fig:mnist5negsampleori}
	\vspace{-0.5cm}
\end{figure}

One meaningful similarity metric for image data is the structural similarity (SSIM) \cite{Wang2004}. For example, Fig.~\ref{fig:mnist5negsampleori} illustrates the division of exemplary classes into typical and atypical normal samples using SSIM. In more detail, an autoencoder is first trained on the whole training set. Subsequently, the similarity metric between reconstructed and original data is calculated. Finally, the data with higher similarity is considered to be typical, whereas the samples with lower similarity are considered to be atypical normal data.

In general, this work is based on the assumption that unknown abnormal data has more common features with the atypical normal data rather than typical normal data in a certain latent space. In particular, we use the atypical normal samples from the given training dataset to model the unknown abnormal classes. This assumption is weaker than those in prior work such as~\cite{Perera2018}, since we do not restrict the unseen abnormal data to a limited number of reference datasets, which may lead to a too loose decision boundary and a subsequent poor sensitivity. Instead, only the intra-class information from the normal class is utilized and a tight decision boundary is thus expected.

In summary, our main contributions in this paper are the following:
\begin{itemize}
	\item \emph{Intra-class splitting} is firstly introduced which is the key to solve feature learning for one-class classification. However, it is a generic method and not limited to one-class feature learning.
	\item A novel intra-class joint training strategy: Both typical and atypical normal subsets have their individual objectives as well as common objectives. These objectives are finally reformulated as a joint optimization problem.
	\item We empirically prove that \emph{closeness} plays a key role in feature learning for one-class problems. Furthermore, it is shown that the combination of \emph{closeness} and \emph{dispersion} can achieve an even better feature extraction performance.
\end{itemize}

\section{Proposed Method}
\label{sec:proposed_method}

\subsection{Overview}
As mentioned in the introduction, the goal is to learn a model which extracts features based on one class only, fulfilling both closeness and dispersion in the latent space.
Therefore, the basic idea is to split the given normal data into typical and atypical normal samples and to use a three-stage joint training procedure.
First, the foundations and strategy of intra-class splitting are explained. Subsequently, three desired characteristics of the latent space are briefly described. Then, the corresponding loss functions are introduced. Finally, an autoencoder-based network considering intra-class dispersion and closeness is presented.

\subsection{Intra-Class Splitting}
In a given normal class, not all samples are representative for this class as illustrated in Fig.~\ref{fig:mnist5negsampleori}. Accordingly, it is assumed that a given normal dataset is composed of two parts, typical normal samples and atypical normal samples. Typical normal samples are the most representative for the normal class and correspond to the majority of the given dataset. In contrast, the remaining samples are considered as atypical normal samples which may mislead the learning of a one-class classifier and thus are used to model the abnormal classes. 

Intuitively, an approach to realize the splitting is utilizing neural networks with a bottleneck structure. By using a compression-decompression process such as in an autoencoder, input data is first transformed into a low-dimensional representation with information loss and then mapped back to the original data space. Hence, only the most important information of the given dataset is well maintained during this process. Accordingly, the samples contain more representative features if they are better reconstructed. 

Formally, a given normal dataset $\chi$ is split by using a predefined similarity metric $f(\bm{x}, \hat{\bm{x}})$ and a ratio $\rho$ where $\hat{\bm{x}}$ is the reconstruction of a sample $\bm{x}\in \chi$. In particular, the first $\rho\%$ samples with the lowest similarity scores are considered as atypical normal samples $\chi_\mathrm{atypical}$, while the others are considered as typical normal samples $\chi_\mathrm{typical}$.

\subsection{Desired Characteristics of the Latent Space}
\label{subsec:characteristics}
\subsubsection{Closeness}
Input data typically distributes along a manifold in the original high-dimensional space. Therefore, it should naturally distribute in a small region in a low-dimensional latent space. In this low-dimensional space, all latent representations of normal data should be as close to each other as possible. In other words, the intra-class latent representations of normal data must have a high closeness among themselves.

\subsubsection{Dispersion}
While closeness is necessary for intra-class latent representations of normal data, abnormal data must be as far away from normal data as possible in the latent space. Hence, abnormal data should have a high dispersion comparing with normal data in this space. As mentioned in the introduction, a training set is assumed to consist of typical and atypical normal data.
As illustrated in Fig.~\ref{fig:ideavenn}, atypical normal samples are forced to distribute far away from typical normal data. Hence, we assume that the unknown abnormal data that behave more like these atypical normal data will also lie far from the normal data.

\subsubsection{Reconstruction information}
While transforming raw data to the latent space, the information contained in high-dimensional data should be retained in order to ensure that the latent space is a compressed representation of the raw data. Thus, we utilize the autoencoder structure as constraints on maintaining high-dimensional information. Equivalent to the ordinary autoencoder, the reconstruction ability of latent representations is quantified by a reconstruction loss.

\subsection{Training}
\label{subsec:training}
As closeness and dispersion are opposite to each other, we designed a joint training procedure for feature learning.
It is performed by three stages using the loss functions defined in Section \ref{subsec:loss_functions}.

\subsubsection{First Stage} All data from the training set is used to train an autoencoder only with a regular reconstruction loss $\mathcal{L}_\mathrm{rec}$. During this stage, the network is considered to be a deep convolutional autoencoder. Note that no constraints are performed on the latent representations at this stage.

\subsubsection{Second Stage} Once the network is trained at the first stage, the similarity between the reconstructed data and the original data is calculated. Afterwards, according to a predefined ratio $\rho$, the first $\rho\%$ of the data with the lowest similarity scores are chosen as the atypical normal samples, whereas the remaining data is typical normal.

\subsubsection{Third Stage} 
To learn proper latent representations, the model is trained with different objectives regarding typical and atypical normal samples, respectively. 

More precisely, the closeness loss $\mathcal{L}_\mathrm{cls}$ acts as the objective for only typical normal samples to force the corresponding latent representations to be close to each other.  In contrast, one dispersion loss $\mathcal{L}_{\mathrm{disp},1}$ is designed for only atypical normal samples to keep them far away from each other in the latent space. Another dispersion loss $\mathcal{L}_{\mathrm{disp},2}$ is taken as an objective by both typical and atypical normal samples to maximize the distances between typical and atypical latent representations. Moreover, a regular reconstruction loss is applied as the objective for all the training samples to retain the high-dimensional information.

The two different training subsets, typical and atypical normal samples, have their own objectives. Nevertheless, all loss terms are reformulated in one unified loss function defined as
\begin{equation}
	\mathcal{L}_\mathrm{rec}+\alpha \mathcal{L}_\mathrm{cls}+\beta_1 \mathcal{L}_{\mathrm{disp},1} + \beta_2 \mathcal{L}_{\mathrm{disp},2}~,
\end{equation}
where $\alpha$, $\beta_1$ and $\beta_2$ balance the ratio between all four loss terms. 

\subsection{Loss Functions}
\label{subsec:loss_functions}

First, the notations used in the loss functions are introduced. 
\begin{itemize}
	\item The tensor $\bm{X}$ denotes a batch of images from the training set. $B$ is the number of images in one minibatch. $\bm{X}_j$ represents the $j$-th image in the given minibatch. $\bm{x}_j$ is the vectorized form of $\bm{X}_j$.
	
	\item The matrix $\bm{Z}$ denotes a batch of latent representations with $L$ being the dimension of the latent vectors. The vector $\bm{z}_j$ is the $j$-th latent representation of $\bm{Z}$.
	
	\item The mapping from the raw data space to the latent space performed by the encoder is denoted as $f_\mathrm{enc}(\cdot)$. Accordingly, the decoder's mapping from the latent space to the original data space is denoted as $f_\mathrm{dec}(\cdot)$. The function $f(\cdot)$ describes a cascade of these two mappings.
\end{itemize}
Based on these notations, we define three loss functions for each of the three characteristics of the latent space described in Section \ref{subsec:characteristics}.

\subsubsection{Reconstruction Loss}
An ordinary mean squared error (MSE) loss is used as reconstruction loss
\begin{equation}
	\label{eq:rec_loss}
	\mathcal{L}_\mathrm{rec} = \frac{1}{B}\sum_{j=1}^{B}||\bm{x}_j - \hat{\bm{x}}_j ||^2~,
\end{equation}
where $\hat{\bm{x}}_j = f_\mathrm{dec}(f_\mathrm{enc}(\bm{x}_j))$ is the reconstruction of $\bm{x}_j $. The minimization of $\mathcal{L}_\mathrm{rec}$ forces the original data to be well reconstructed from the latent space. Correspondingly, the high-dimensional information is retained during the training. Moreover, this term helps to avoid trivial solutions, e.g. all-zero latent representations.

\subsubsection{Closeness Loss}
The closeness requirement states that each typical normal latent representation $\bm{z}_j$ should have a small distance to any another randomly chosen typical normal latent representation $\bm{z}_{i\ne j}$, where arbitrary distance metrics can be used for this purpose. Specifically, it is a metric based on the Euclidean distance
\begin{equation}
	\label{eq:cls_loss}
	\mathcal{L}_\mathrm{cls}=\frac{1}{B}\sum_{j=1}^{B}\sqrt{\frac{1}{L}||\bm{z}_j-\bm{z}_{i\ne j}||^2}~,
\end{equation}
with $\bm{z}_j = f_\mathrm{enc}(\bm{X}_j)$ and $\bm{z}_{i\ne j} = f_\mathrm{enc}(\bm{X}_{i\ne j})$. This minimization has the same expectation as if we calculated all distances from other latent representations, since the mini-batch stochastic gradient decent method is used in this work.

\begin{figure*}
	\centering
	\includegraphics[scale=0.6]{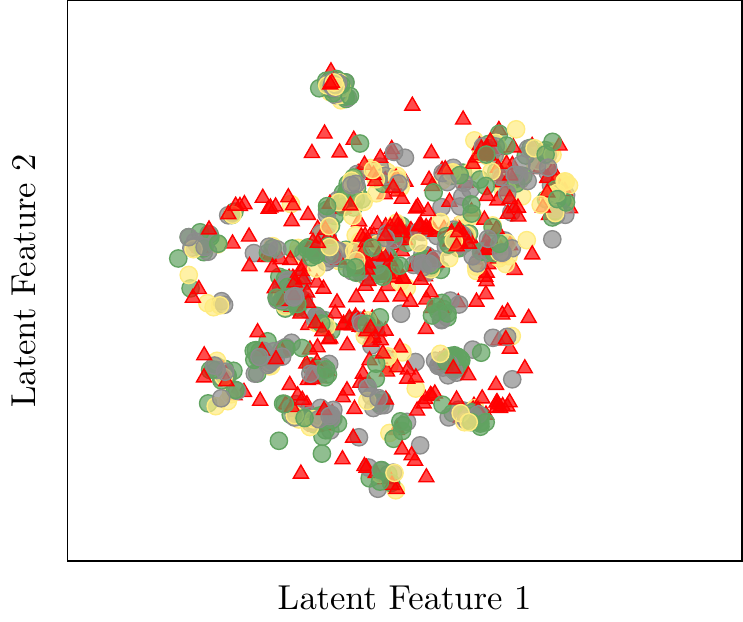}
	\hspace{1cm}
	\includegraphics[scale=0.6]{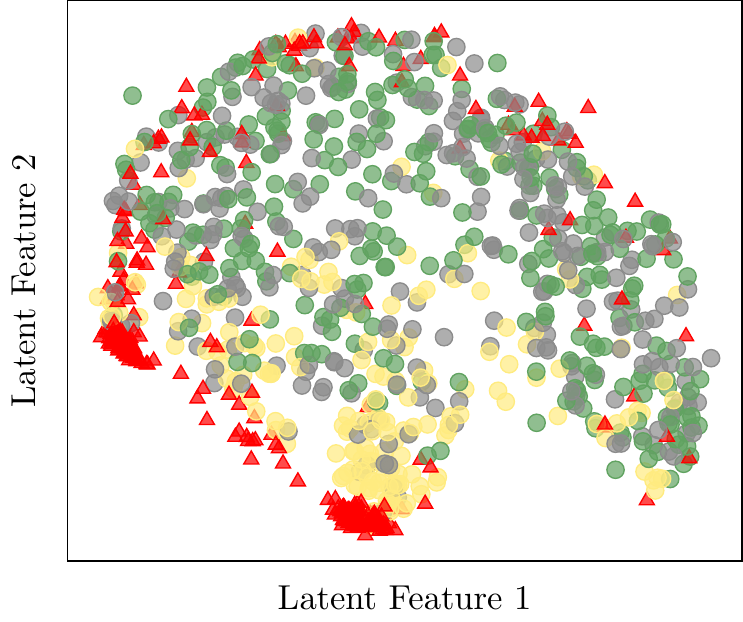}
	\hspace{1cm}
	\raisebox{1.2cm}{\includegraphics[scale=0.8, trim={1.8cm 0cm 1.2cm 6.5cm}, clip]{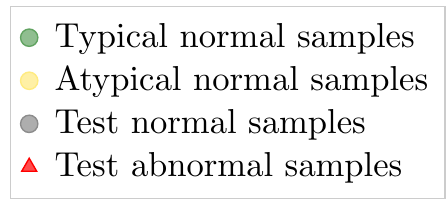}}
	\caption[Latent representation]{If only trained with the closeness loss, the network tends to learn a too simple function to map all samples to a small region \emph{(left)}. Accordingly, the normal data \emph{(green, yellow and gray)} and abnormal data \emph{(red)} distribute in an indistinguishable mixture. However, with the dispersion loss \emph{(right)}, the typical normal samples \emph{(green)} distribute compactly, while the atypical normal samples \emph{(yellow)} distribute far away or around the typical normal ones. Finally, the abnormal \emph{(red)} samples are thus also far away from the normal ones.}
	\label{fig:latent_comp}
\end{figure*}

The above loss function alone has the tendency to map all raw data, both from the normal class and the abnormal classes, to a small region in the latent space. Fig.~\ref{fig:latent_comp} shows this effect for the dataset Fashion-MNIST. As a result, latent representations of normal and abnormal data distribute in a mixture. 

\subsubsection{Dispersion Loss}
The dispersion loss between these typical and atypical normal subsets is defined as
\begin{equation}
	\label{eq:disp_loss}
	\begin{split}
		\mathcal{L}_\mathrm{disp} = &\beta_1\cdot
		\underbrace{\Big(-\frac{1}{B}\sum_{j=1}^{B}\sqrt{\frac{1}{L}||\bm{z}_{j,\mathrm{atypical}}-\bm{z}_{i\ne j,\mathrm{atypical}}||^2}\ \Big)}_{\mathcal{L}_{\mathrm{disp},1}} \\ 
		&+ \beta_2\cdot \underbrace{\Big(-\frac{1}{B}\sum_{j=1}^{B}\sqrt{\frac{1}{L}||\bm{z}_{j,\mathrm{atypical}}-\bm{z}_{j,\mathrm{typical}}||^2}\ \Big)}_{\mathcal{L}_{\mathrm{disp},2}}~,
	\end{split}
\end{equation}
where $\bm{z}_{j,\mathrm{atypical}}=f_\mathrm{enc}(\bm{X}_{j,\mathrm{atypical}})$ and $\bm{z}_{j,\mathrm{typical}}=f_\mathrm{enc}(\bm{X}_{j,\mathrm{typical}})$. $\bm{X}_{j,\mathrm{typical}}$ denotes randomly chosen typical normal samples. The minimization of $\mathcal{L}_{\mathrm{disp},1}$ forces latent representations of atypical normal data to be far away from each other which results in a dispersion among atypical normal latent representations. Moreover, minimizing $\mathcal{L}_{\mathrm{disp},2}$ leads to large distances between typical and atypical normal samples.

\subsection{Network Architecture}
\label{sec:network_architecture}
The proposed feature learning method is modeled by a deep convolutional autoencoder based on AlexNet~\cite{krizhevsky2012imagenet}. Both encoder and decoder are used during the training procedure. Once the model is trained, only the encoder part is utilized as a feature extractor.

\section{Experiments}
\label{sec:experiments}
The proposed method was used to extract latent features of different datasets which were subsequently fed into a one-class classifier. OCSVMs achieved state-of-the-art performance on extracted features, so an OCSVM was used to perform one-class classification.
In one experiment, images of one class were selected as normal data, while the images of all remaining classes were considered as abnormal data and were not available during training. Each experiment was repeated five times with different initializations.

\begin{table*}[t]
	\centering
	\caption{Balanced Accuracy in \%.}
	\label{tab:baccu}
	\vspace{-0.2cm}
	\def\arraystretch{1.15}
	\begin{tabular}{lccccccc}
		\toprule
		\textbf{Dataset} & \textbf{HOG} & \textbf{PCA}	& \textbf{ImageNet}	& \textbf{CAE}	& \textbf{Original}	& \textbf{CLS} & \textbf{Ours}\\
		\midrule
		MNIST & 64.3 $\pm$0.0 & 75.2 $\pm$0.0 & 68.7 $\pm$0.0 & 85.2 $\pm$0.7 & 84.2 $\pm$0.0 & 84.7 $\pm$5.6 & \textbf{91.3 $\pm$0.7 } \\
		FMNIST & 81.1  $\pm$0.0 	& 81.7 $\pm$0.0 & 65.2 $\pm$0.0	& 81.6 $\pm$1.0 & 82.3 $\pm$0.0 & 84.8 $\pm$2.5 	& \textbf{87.7 $\pm$0.5 } \\
		CIFAR-10 & 53.8 $\pm$0.0 & 57.1 $\pm$0.0 & 53.7 $\pm$0.0 & 56.9 $\pm$1.4 & 56.5 $\pm$0.0 & 54.6 $\pm$3.0 & \textbf{60.6 $\pm$1.2 } \\
		\bottomrule
	\end{tabular}
\vspace{-0.2cm}
\end{table*}

\subsection{Experimental Setup}
\label{sec:experimental_setup}

\subsubsection{Datasets}
The proposed method was evaluated on the datasets MNIST~\cite{LeCun1998}, Fashion-MNIST (FMNIST)~\cite{Xiao2017} and CIFAR-10~\cite{cifar10}. All three datasets are composed of 10 different classes. The number of training data for all experiments was set to 4000. The numbers of normal and abnormal samples in the test set were 1000 and 9000, respectively\footnote{The MNIST dataset has about but not exactly 1000 normal and 9000 abnormal samples for each individual experiment.}. For example, in one individual experiment, if digit 2 from dataset MNIST was considered as normal data, then the training set consisted of 4000 different images of digit 2. Furthermore, the test set consisted of 9000 other images of digit 2 and 1000 images of the other nine digits from 0 to 9 except of digit 2.\footnote{There are 20\% samples of the test set randomly selected as validation set to choose a proper classification threshold for the metric balanced accuracy.}

Before training, all image pixels were normalized to the range $[0, 1]$ by min-max scaling. Note that images from MNIST and Fashion-MNIST have the size of $28 \times 28$, while images from CIFAR-10 have the size of $32\times 32 \times 3$.

\subsubsection{Hyperparameters}
The proposed model was implemented with TensorFlow and Keras~\cite{abadi2016tensorflow}. L2-regularization was used for every convolutional layer with a regularization parameter of $10^{-6}$. The training minibatch size was 64 for all experiments. The ratio $\rho$ for choosing atypical normal samples was set to 10, unless otherwise stated. $\alpha$, $\beta_1$ and $\beta_2$ defined in Section \ref{subsec:loss_functions} were chosen to be 1, $10^{-5}$ and $10^{-5}$, unless otherwise specified. The dimension of the latent space was set to 64. The one-class classifier used in this work was the one-class support vector machine (OCSVM) with $\nu=0.1$ and $\gamma=\frac{1}{\#\mathrm{features}}$.

\subsubsection{Baseline Models}
To the best of our knowledge, few prior works were designed for image-level one-class feature extraction. Therefore, the following conventional feature extraction methods were used as shallow baseline models:
\begin{itemize}
	\item \emph{Original features (Original):} The original images were vectorized as the input for OCSVM. Correspondingly, samples from MNIST and  FMNIST have 784 features and those from CIFAR-10 have 3072 features.
	\item \emph{Principle Component Analysis (PCA):} The first 64 PCA components of each input image were used as the input for OCSVM.
	\item \emph{Histogram of oriented gradients (HOG):} HOG features~\cite{Dalal2005} for each sample were extracted as the input for OCSVM. The length of HOG features for the samples from MNIST and FMNIST was 144 and for CIFAR-10 was 324.
\end{itemize}
Furthermore, the following deep feature extraction methods were considered as baselines in this work:
\begin{itemize}
	\item \emph{Pretrained Features (ImageNet):} Features extracted by a VGG19~\cite{simonyan2014very} pretrained on ImageNet~\cite{russakovsky2015imagenet} were used as the input for OCSVM.
	\item \emph{Convolutional Autoencoder (CAE):} Features extracted by a regular autoencoder without any constraints on the latent space were used for subsequent one-class classification. 
	\item \emph{Autoencoder with closeness loss (CLS):} An autoencoder was trained with the proposed closeness loss as a regularization but without intra-class splitting to extract features for one-class classification.
\end{itemize}
Note that both the baselines CAE and CLS shared the same architecture with the proposed method.

\subsubsection{Metrics}
In one-class classification, imbalanced data is common. Hence, we used balanced accuracy as metric, because it allows a fair comparison between our method and the baseline models also for imbalanced datasets. The balanced accuracy is defined as
\begin{equation}
	\label{baccu}
	\mathrm{BACC} = \frac{1}{2}\Big(\frac{\mathrm{TP}}{\mathrm{TP}+\mathrm{FN}}+\frac{\mathrm{TN}}{\mathrm{TN}+\mathrm{FP}}\Big)~.
\end{equation}
with the terms true positive (TP), true negative (TN), false negative (FN) and false positive (FP). In this work, the normal class (the known class during training) was considered as the negative class. In contrast, the abnormal classes (unknown classes during training) were considered as positive classes.

\subsection{Results and Discussion}
\label{subsec:results_and_discussion}
Table~\ref{tab:baccu} shows the results averaged over all classes. Per class, the performance was determined by the mean and standard deviation of the balanced accuracies for different initializations.
The results indicate that the proposed method performed best compared to all baseline models on all datasets. In particular, our method significantly outperformed the deep baseline models, i.e. the ImageNet and CAE features. Although CLS, a regular autoencoder only with the proposed closeness regularization, achieved a good performance, adding an additional dispersion constraint further improved the performance in all experiments.

\begin{figure}
	\centering
	\includegraphics[width=\linewidth]{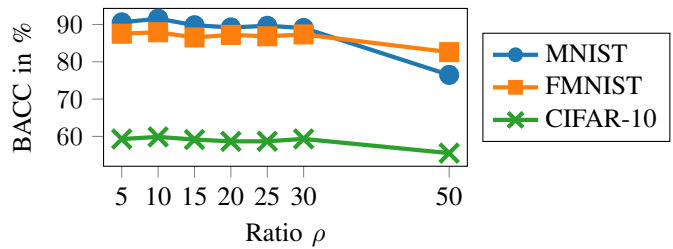}
	\vspace{-0.5cm}
	\caption[ratio]{Balanced accuracy vs. ratio $\rho$.}
	\label{fig:ratio}
\end{figure}

Fig.~\ref{fig:ratio} shows the balanced accuracies averaged over classes in relation to different ratios $\rho$. In general, the proposed method was not sensitive to the choice of the ratio $\rho$ for splitting the training data. However, the balanced accuracies tended to be lower if the ratio was higher, e.g. $\rho=50$, because a larger ratio indicates that more samples are considered to be atypical normal. This will lead to a low true negative rate.

\begin{figure}
	\centering
	\includegraphics[width=\linewidth]{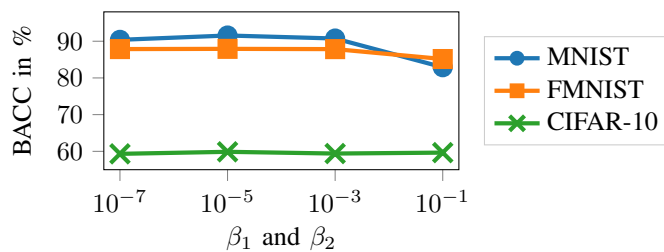}
	\vspace{-0.5cm}
	\caption[beta]{Balanced accuracy vs. $\beta_1$ and $\beta_2$.}
	\label{fig:beta}
\end{figure}

Fig.~\ref{fig:beta} illustrates the balanced accuracies averaged over classes in relation to different values of $\beta_1$ and $\beta_2$. In the range of $10^{-7}$ to $10^{-3}$, the performance of the proposed method was stable for the conducted experiments. In contrast, the balanced accuracies on MNIST and FMNIST decreased with $\beta_1 = \beta_2 = 10^{-1}$, because the dispersion loss term is dominant. This is not desirable for an OCSVM.
\balance

\section{Conclusion}
\label{sec:conclusion}
In this work, we presented a novel generic feature learning method for one-class classification. To learn a proper latent space, normal training samples were split into typical and atypical normal data. These two subsets were used to train a network with different losses in an joint way under the constraints of the proposed closeness and dispersion requirements. As a result, the trained feature extractor enabled to extract highly discriminative features between normal and abnormal data.
Various and intensive experiments on three image datasets used the extracted features as the input for an OCSVM to perform one-class classification. In all conducted experiments, the method outperformed all other baseline models. In particular, our method showed a large improvement over ImageNet features, CAE and the original features.
An ongoing work is to realize the proposed method in an end-to-end way since it has already shown its efficient feature extraction ability for one-class classification.

\bibliographystyle{IEEEtran}
\bibliography{references}
\end{document}